\documentclass[sigconf,10pt,nonacm]{acmart}
\AtBeginDocument{}

\setcopyright{acmlicensed}
\copyrightyear{2018}
\acmYear{2018}
\acmDOI{XXXXXXX.XXXXXXX}
\acmConference[HotMobile'26]{The 27th International Workshop on Mobile Computing Systems and Applications}{February 25--26,
  2026}{Atlanta, GA}
\acmISBN{978-1-4503-XXXX-X/2018/06}

\acmSubmissionID{97}

\usepackage{hyperref}
\hypersetup{
colorlinks,
linkcolor=blue,
urlcolor=blue,
}

\usepackage{xcolor}
\usepackage{xspace}
\usepackage{balance}
\usepackage{amsmath}
\usepackage{multirow}
\usepackage{colortbl}
\usepackage{lipsum}
\usepackage{caption}
\usepackage[labelformat = parens, labelsep = space, textfont=small]{subcaption}

\usepackage{enumitem}
\usepackage[most]{tcolorbox} \usepackage{arydshln}
\usepackage{tabularx}
\usepackage{etoolbox}

\definecolor{pro_green}{rgb}{0.0, 0.66, 0.47}
\definecolor{overleaf_green}{rgb}{0.08, 0.54, 0.02}

\newcommand{\1}{{\em (i)}}
\newcommand{\2}{{\em (ii)}}
\newcommand{\3}{{\em (iii)}}

\newcommand{\para }[1]{\noindent  {\bf #1}}
\setlist[itemize]{leftmargin=.12in}

\definecolor{lipsumcolor}{RGB}{0,70,180}

\begin{document}

\title[Can Foundation Models Revolutionize Mobile AR Sparse Sensing?]{Can Foundation Models Revolutionize \\ Mobile AR Sparse Sensing?}

\author{Yiqin Zhao}
\email{yzigm@rit.edu}
\orcid{0000-0003-1044-4732}
\affiliation{\institution{Rochester Institute of Technology}
  \city{Rochester}
  \state{NY}
  \country{USA}
}

\author{Tian Guo}
\email{tian@wpi.edu}
\orcid{0000-0003-0060-2266}
\affiliation{\institution{Worcester Polytechnic Institute}
  \city{Worcester}
  \state{MA}
  \country{USA}
}

\renewcommand{\shortauthors}{Zhao and Guo}

\begin{abstract}
Mobile sensing systems have long faced a fundamental trade-off between sensing quality and efficiency due to constraints in computation, power, and other limitations. Sparse sensing, which aims to acquire and process only a subset of sensor data, has been a key strategy for maintaining performance under such constraints. However, existing sparse sensing methods often suffer from reduced accuracy, as missing information across space and time introduces uncertainty into many sensing systems.
In this work, we investigate whether foundation models can change the landscape of mobile sparse sensing. Using real-world mobile AR data, our evaluations demonstrate that foundation models offer significant improvements in geometry-aware image warping, a central technique for enabling accurate reuse of cross-frame information. Furthermore, our study demonstrates the scalability of foundation model-based sparse sensing and shows its leading performance in 3D scene reconstruction. Collectively, our study reveals critical aspects of the promises and the open challenges of integrating foundation models into mobile sparse sensing systems.
\end{abstract}

\ccsdesc[500]{Computing methodologies~Mixed / augmented reality}
\ccsdesc[500]{Human-centered computing~Ubiquitous and mobile computing systems and tools}

\keywords{Mobile AR, Mobile Sensing, Sparse Sensing, Foundation Model, Depth Estimation}

\maketitle

\section{Introduction}
\label{sec:introduction}

Today's mobile systems employ sensing systems to understand their surrounding environments to enable rich user experiences. In many mobile sensing systems, including our focus of augmented reality, deep learning models have been a central technology that provides accurate mobile sensing on complex sensor data.
However, continuously sensing, often necessary to provide required features for mobile AR applications, can pose a high toll on mobile energy.

One natural way is to sense less, which we generally term \emph{sparse sensing}, but still be able to deliver the same or similar application performance.
Traditional sparse sensing techniques reduce system overhead by selectively activating subsets of sensors or processing pipelines~\cite{yang2015adaptive}. However, such systems often struggle to maintain real-world robustness.

The advent of foundation models could be a game-changer for sparse sensing. By leveraging their large-scale pretraining and strong generalization capabilities, these models exhibit remarkable robustness in extracting meaningful information even from sparse inputs. For example, an image foundation model can infer omnidirectional environment lighting from just one or a few camera frames~\cite{zhao2025clear}. Moreover, recent models, such as DINOv3~\cite{simeoni2025dinov3}, are capable of producing high-resolution sensing results directly from RGB images with a quality that was previously unattainable even with advanced sensor hardware.

In this work, we set out to explore the promises and open challenges when leveraging foundation models for mobile sparse sensing. We focus on mobile augmented reality, which is a fast-growing mobile computing area that deeply depends on multimodal sensing.
Specifically, we investigate two research questions:
\1 The feasibility of leveraging foundation models on sparse sensing with a data-driven evaluation to quantify the improvement of cross-frame information reuse.
\2 The scalability of sparse sensing on long-duration AR sessions by quantifying the impacts on 3D reconstruction, an important downstream task.

To answer the abovementioned research questions, we use an indoor dataset called Scannet++~\cite{yeshwanth2023scannet++} which consists of real-world mobile AR data recordings. We begin by investigating how foundation models can be leveraged to assist geometry-aware image warping, a central technique for supporting cross-frame information reuse. We tested geometry-aware image warping across different frame intervals, representing different sparsity levels in the temporal domain. In the experiment, we used the device LiDAR depth, the foundation model estimated depth, and the ground truth depth for warping. We measure the error using SSIM on warped RGB and depth images. Our results show that foundation model-based warping significantly outperforms LiDAR-based one, with an average improvement of at least 25.5\%.

We also investigate the scalability of sparse sensing on long-duration AR sessions and find that foundation model-based 3D reconstruction quality, even with aggressive temporal downsampling, significantly outperforms LiDAR-based reconstruction with 60 FPS sensing. Specifically, using Hausdorff Distance as the metric, foundation model-based reconstruction outperforms LiDAR-based one by 48\%. This indicates the possibility of using sparsely sensed data for longer-term environment understanding.

Finally, we evaluate the temporal and spatial information differences in AR sessions. We quantify frame-to-frame information overlap under different sparse sensing policies. Our results show that, on average, only about 27\% of frames in an AR session are needed to achieve $>=$ 80\% information overlap between all consecutive frames. However, none of the traditionally used time interval-based or motion-based control policies can achieve comparable performance. This leaves open questions and challenges for future sparse sensing control policy design.
Our observations also point toward a new type of sparse sensing policy design that considers both temporal and spatial dimensions.

In summary, we make the following key contributions:

\begin{itemize}[leftmargin=*,topsep=0pt,noitemsep]
\item We demonstrate the opportunities to apply sparse sensing in mobile AR with an analysis of how different depth estimation methods impact geometry-aware image warping. We show that foundation models can substantially enhance the accuracy of reusing information in real-time AR sessions.
\item We show the feasibility of leveraging foundation models to improve the information reuse accuracy on temporally adjacent frames, which could, in turn, allow the use of sparser sensing compared to not using foundation models.
\item We show that with the help of foundation models, we can achieve comparable or even better 3D reconstruction quality over long-duration AR sessions when compared to sensing with a much higher frequency with a LiDAR sensor, i.e., 15FPS vs. 60FPS.
\end{itemize}

\section{Background}
\label{sec:background}

\para{Sparse sensing.}
Sparse sensing aims to achieve accurate sensing from sensor data that contains limited information. Traditional sparse-sensing research often focuses on inferring information from compressed, undersampled, or partial signals to reconstruct rich results from fewer measurements~\cite{donoho2006compressed}. On mobile platforms, sparse sensing is often used as a strategy to reduce mobile system resource usage by using only a small subset of sensor measurements or computation resources. Existing systems often employ sparse sensing by carefully designing control algorithms and systems to selectively activate sensors or processing frames based on task importance or environmental dynamics~\cite{razzaque2014energy,chen2015dass}. However, sparse inputs often lead to degraded estimation quality since many vision or sensing algorithms rely on dense, temporally consistent data. To address this, prior work~\cite{yang2015adaptive,duarte2009learning} explores adaptive sampling and predictive sensing to balance efficiency and accuracy, yet these approaches remain limited by their task-specific heuristics.

\para{Foundation model.}
Foundation models are large-sized and multi-task-capable models pretrained on diverse datasets. They offer strong generalization and robustness for several tasks.  The emergence of foundation models has transformed tasks across computer vision, natural language processing, and robotics. In vision, image encoders such as DINOv3~\cite{simeoni2025dinov3} and SAM~\cite{ravi2024sam} demonstrate strong zero-shot segmentation and object recognition capabilities. In multimodal learning, models like FLAVA~\cite{singh2022flava} and Florence~\cite{yuan2021florence} integrate text, vision, and geometry, enabling joint reasoning over heterogeneous sensory inputs. Unlike traditional task-specific models that are prone to overfitting, foundation models learn rich, transferable representations that capture both semantic and structural relationships across data distributions. However, integrating foundation models into mobile sensing still remains challenging~\cite{yuan2024mobile}. Most foundation models contain billions of parameters, leading to high computational costs and substantial energy consumption.

\section{Experiment Setups}
\label{sec:experiment_setups}

\para{Study aims}.
To explore the promises and challenges of leveraging foundation models to sparse sensing, we design three experiments: cross-frame information reuse (\S\ref{sec:cross_frame_information_reuse}), long-duration sparse sensing (\S\ref{sec:long_duration_sparse_sensing}), and spatial-temporal sparse sensing (\S\ref{sec:spatial_temporal_sparse_sensing}). Specifically, the first experiment examines the feasibility of foundation models by demonstrating frame-level perception results via geometry-aware image warping. The second experiment focuses on understanding foundation models' ability in improving long-duration sparse sensing with a downstream task called 3D reconstruction. The last experiment analyzes the information overlap under both temporal and spatial domains, paving new directions to perform sparse sensing in the era of foundation models.

\begin{figure}[t]
\centering
    \includegraphics[width=\linewidth]{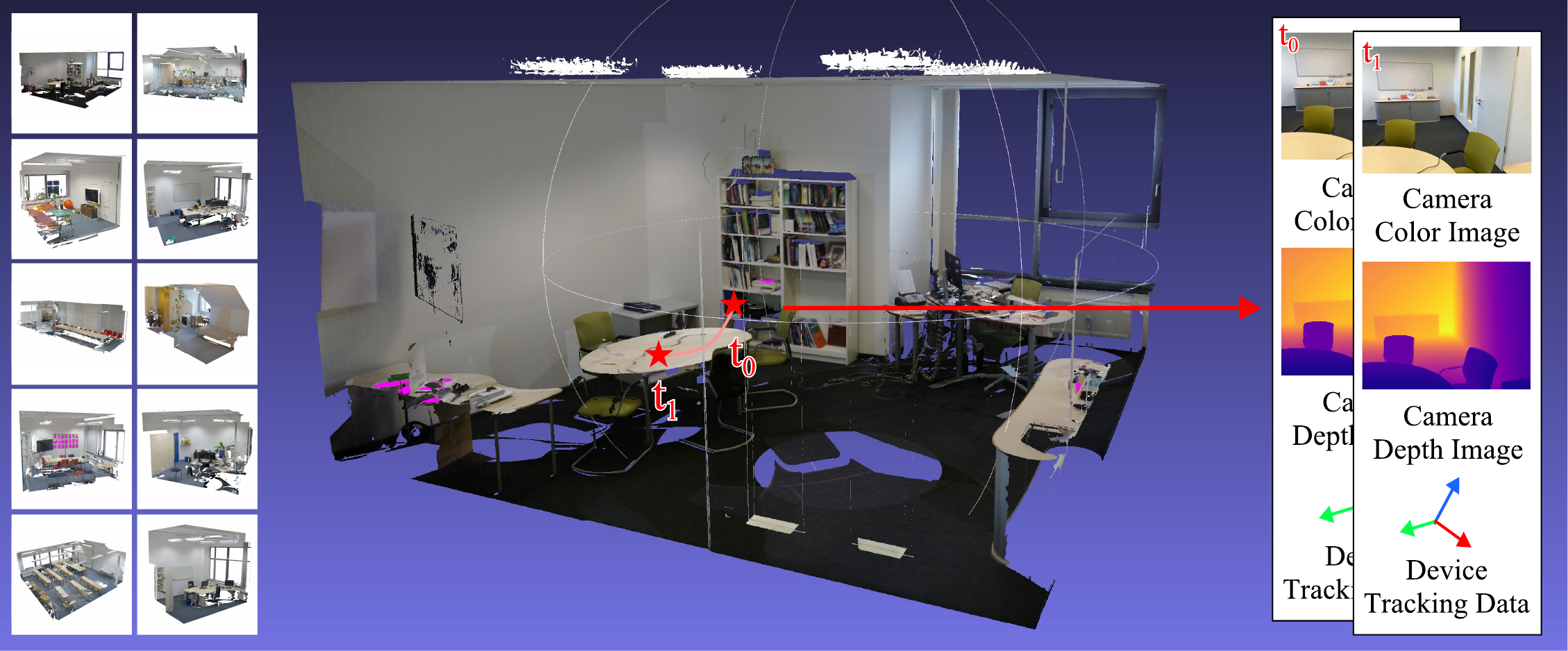}
    \caption{
        Experiment environment setup.
        \textnormal{We utilize ScanNet++~\cite{yeshwanth2023scannet++}, a state-of-the-art high-quality 3D indoor scene reconstruction dataset, to build our experiment environment. From the dataset, we extract iPhone-based AR session recordings with real-world device mobility and sensor data, as well as laser-scanner-based 3D reconstruction geometries that provide environment sensing ground truth.
}
    }
    \label{fig:exp_env_setup}
\end{figure}
\para{Dataset setup}. Our study focuses on two aspects of mobile sparse sensing: cross-frame information reuse (\S\ref{sec:cross_frame_information_reuse}) and spatial-temporal sparse sensing (\S\ref{sec:spatial_temporal_sparse_sensing}). To explore key questions in these areas, we utilize real-world 3D scans and mobile sensor data from ScanNet++~\cite{yeshwanth2023scannet++}. Figure~\ref{fig:exp_env_setup} shows a set of 3D scene examples of the ScanNet++ dataset and samples of our extracted data frames. Each extracted data frame contains an iPhone camera RGB image, an iPhone LiDAR depth image, and an ARKit pose-tracking result. Using the 3D environment scans, we also extract a ground-truth depth map from the scanned geometry, which is generated by precise laser scanners. Each scene contains about 10,000 data frames, with a framerate of 60. In total, we extract 10 3D scenes and 1,500 minutes of mobile AR data frames.

\para{Implementation.}
Our experiment tools and systems are primarily implemented in Python. We leverage the \texttt{diffusers}\footnote{Diffusers: \url{https://huggingface.co/docs/diffusers/index}} and \texttt{transformers}\footnote{Transformers: \url{https://huggingface.co/docs/transformers/en/index}} libraries to perform inference with foundation models. For depth estimation in (\S\ref{sec:cross_frame_information_reuse}), we employ the geometry understanding foundation model \texttt{Metric3DV2}~\cite{hu2024metric3dv2} with the \texttt{metric3d\_vit\_large} checkpoint. Rendering is implemented using \texttt{ModernGL}\footnote{ModernGL: \url{https://github.com/moderngl/moderngl}}, which provides a Python interface to the standard OpenGL graphics pipeline. In (\S\ref{sec:cross_frame_information_reuse}), we adopt the screen-space meshing technique~\cite{Du2020DepthLab} to enable rasterization-based vertex interpolation during image warping. To prevent geometry artifacts, triangles with areas exceeding the 95th percentile of triangle areas are discarded. For 3D reconstruction experiments, we use the Open3D\footnote{Open3D: \url{https://open3d.org}} framework for mesh-related processing. All experiments are conducted on an NVIDIA GH200 Grace Hopper-based platform equipped with 64 ARM CPU cores, 432~GiB of system memory, and 96~GiB of GPU memory.

\para{Evaluation metrics.}
For the cross-frame information reuse experiment (\S\ref{sec:cross_frame_information_reuse}), we evaluate the effectiveness of geometry-aware image warping by measuring the accuracy of the warped RGB and depth pixel values. Specifically, we employ the structural similarity index (SSIM)~\cite{wang2004image} to assess the structural differences between warped RGBD images.
For assessing the quality of 3D reconstruction in (\S\ref{sec:long_duration_sparse_sensing}), we use the Hausdorff Distance~\cite{huttenlocher2002comparing} on the mesh vertices.
For the spatial-temporal sparse sensing experiment (\S\ref{sec:spatial_temporal_sparse_sensing}), we assess the camera pose differences using $\text{SE}(3)$ geodesic distance following~\cite{triggs1999bundle,gao2023fully,govindu2004lie}.

\section{Cross-Frame Information Reuse}
\label{sec:cross_frame_information_reuse}

Sparse sensing is often enabled by reusing sensing results across multiple frames~\cite{kong2023accumo,meng2021edgear,li2020towards}. However, a central challenge in cross-frame reuse lies in accurately transforming information between frames of different view poses. In this section, we explore how foundation models can be leveraged to address the key challenges of cross-view information transformation. Also, we measure the quantitative quality impacts on cross-frame information reuse.

\para{Geometry-aware image warping} is a widely adopted technique for enabling mobile sensing at reduced framerates by reusing sensing results across temporally adjacent frames~\cite{kong2023accumo,meng2021edgear}.
It allows cross-view information sharing over overlapping 3D regions by transferring pixel-level information from one view to another based on the underlying environment geometry.
Formally, given a pixel $\mathbf{p}_t = [u_t, v_t, 1]^\top$ in frame $t$ at $u, v$ with depth $D_t(\mathbf{p}_t)$, camera intrinsics $\mathbf{K}$, and relative rotation and translation $(\mathbf{R}_{t \to t'}, \mathbf{t}_{t \to t'})$ between frame $t$ and $t'$, its corresponding pixel $\mathbf{p}_{t'} = [u_{t'}, v_{t'}, 1]^\top$ in the target frame can be obtained by
\begin{equation}
\mathbf{p}_{t'} \sim \mathbf{K} \left( \mathbf{R}_{t \to t'} \, D_t(\mathbf{p}_t) \, \mathbf{K}^{-1} \mathbf{p}_t + \mathbf{t}_{t \to t'} \right),
\end{equation}
where $\sim$ denotes equality up to a homogeneous scaling factor.
This warping process allows per-pixel attributes, such as color, depth, or semantic labels, from frame $t$ to be geometrically aligned and transferred to frame $t'$, thereby enabling sparse mobile sensing systems to maintain temporal consistency even under reduced sensing frequency.

This transformation process relies on the accurate understanding of the depth $D_t(\mathbf{p}_t)$. Prior mobile depth estimation methods have often failed to achieve accurate depth estimation results on complex real-world environment geometries, even with specialized hardware like the LiDAR sensor. However, recent foundation model-based depth estimation methods have shown significant improvement in the accuracy of depth estimation as well as the quality of understanding fine-grained details, thanks to the rich image understanding prior of these models.

\para{Accuracy evaluation}.
Next, we evaluate the quantitative accuracy of geometry-aware image warping with different camera depth data.
For this experiment, we first select paired AR frame samples from ScanNet++ by randomly choosing frame pairs within a temporal window of $[10, 100]$ frames, corresponding to $[160, 1600]$ milliseconds of time difference between frames. We use a step of 10 to find frames within the time window.
This random pairing simulates view pose variations induced by user mobility in real-world AR usage. Using these paired frames, we then perform geometry-aware image warping under all three different depth inputs: \1 LiDAR depth, \2 foundation model–predicted depth, and \3 ground truth depth.

\begin{figure}[t]
\centering
    \includegraphics[width=\linewidth]{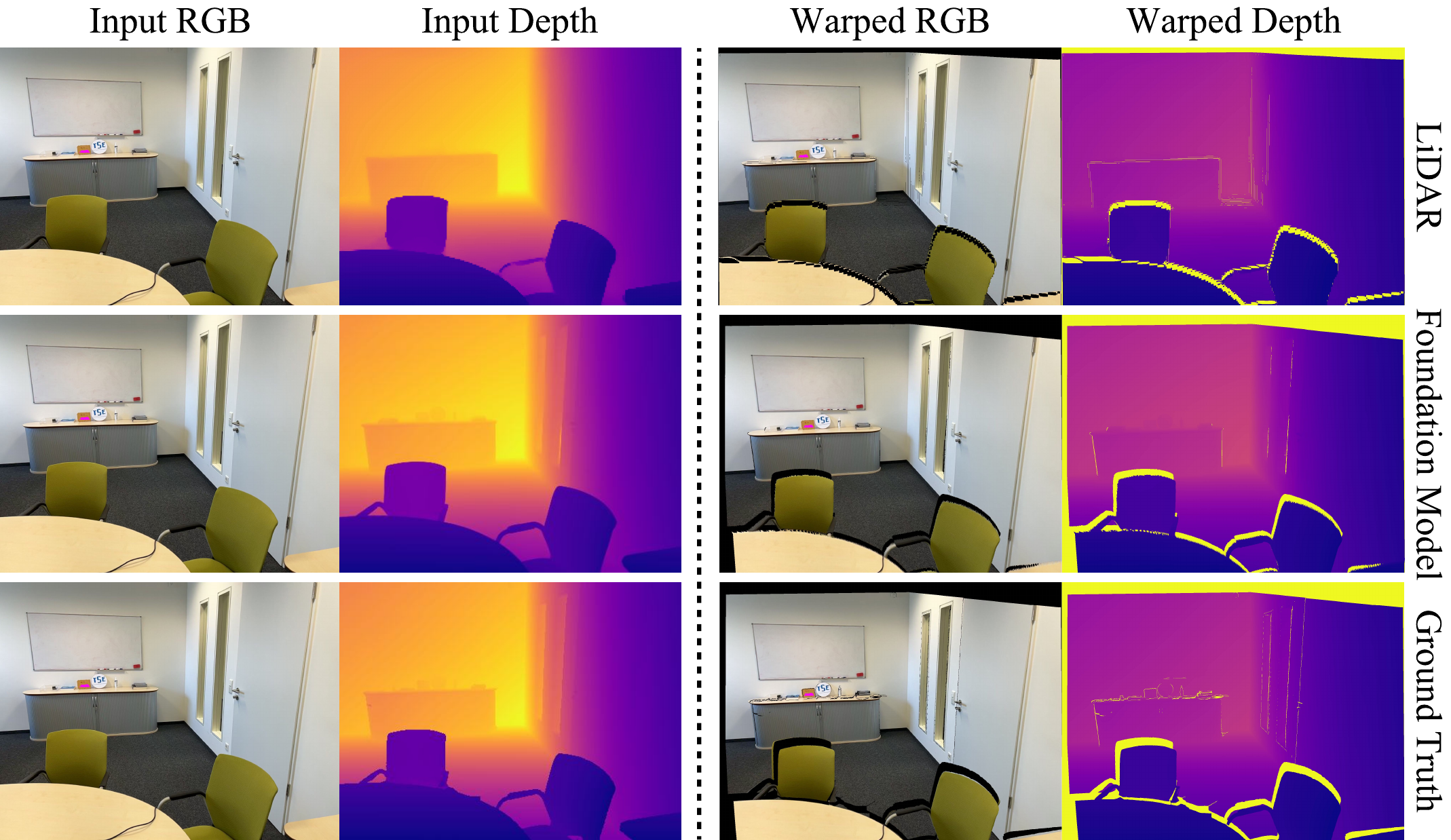}
    \caption{
        Qualitative comparison on geometry-aware image warping.
        \textnormal{We show comparisons on geometry-aware image warping with LiDAR depth, foundation model estimated depth, and ScanNet++ ground truth depth. The time difference between the warping source and the target is 10 frames. We observe that high-quality depth map details estimated by foundation models significantly improve image warping accuracy.}
    }
    \label{fig:qualitative_comparision_geometry_aware_image_warping}
\end{figure}

In Figure~\ref{fig:qualitative_comparision_geometry_aware_image_warping}, we show a qualitative comparison of geometry-aware image warping using depth from foundation models, a LiDAR sensor, and the ground truth depth. For our experiment, we assume the environment does not change between views. We notice that not only does the foundation model generate more fine-grained depth details, but it also gives more accurate results on image edges and object boundaries. The improved depth quality translates to fewer visual artifacts and more accurate results on image warping. Consequently, the warping accuracy can enable more accurate reuse of sparse sensing results in real-time mobile AR.

\begin{figure}[t]
\centering
    \begin{minipage}{0.6\linewidth}
        \centering
        \includegraphics[width=\linewidth]{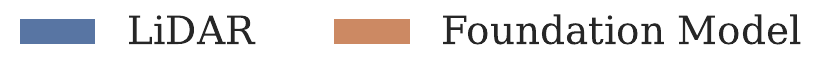}
    \end{minipage}
    \vspace{.25em}

    \begin{subfigure}[b]{0.485\columnwidth}
        \centering
        \includegraphics[width=\linewidth]{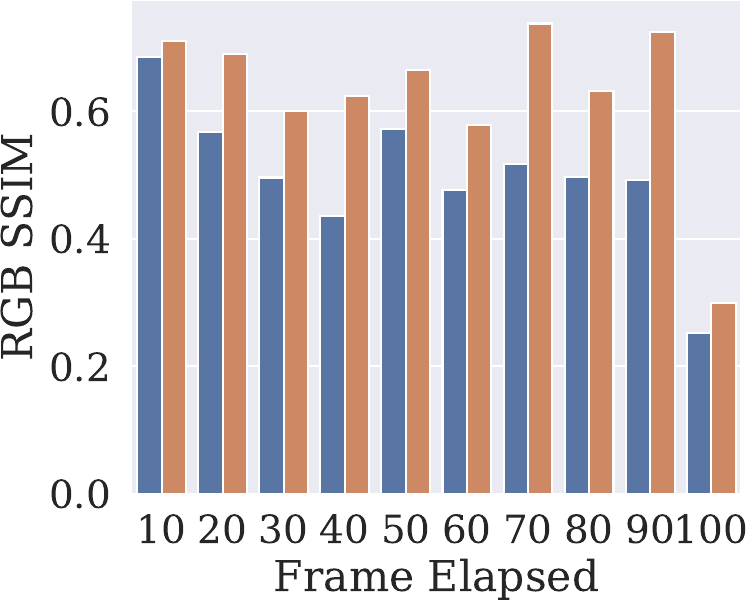}
        \caption{RGB SSIM$\uparrow$}
        \label{subfig:warp_acc_rgb}
    \end{subfigure}\hfill
    \begin{subfigure}[b]{0.485\columnwidth}
        \centering
        \includegraphics[width=\linewidth]{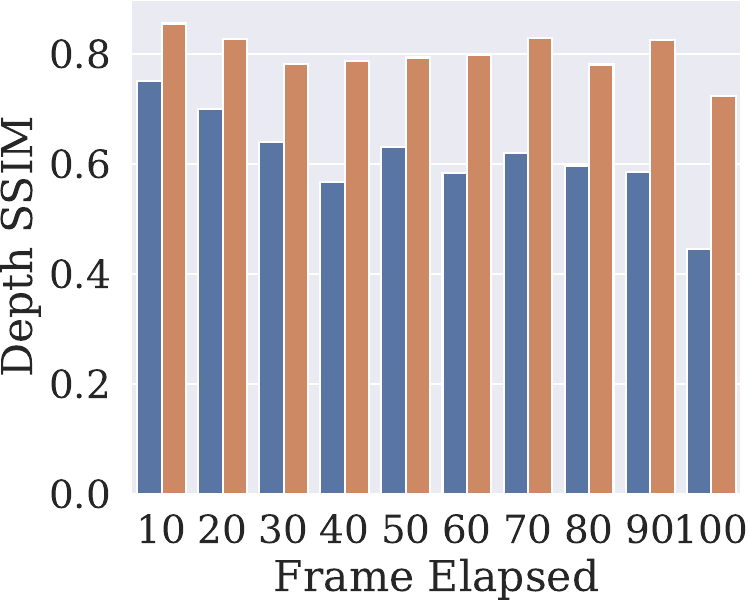}
        \caption{Depth SSIM$\uparrow$}
        \label{subfig:warp_acc_depth}
    \end{subfigure}
    \caption{
        Cross-frame information reuse accuracy.
        \textnormal{For both warped RGB and depth images, image warping based on foundation model–estimated depth consistently yields higher SSIM values. Moreover, the foundation model–based warping demonstrates greater robustness under larger temporal gaps between frames.}
    }
    \label{fig:quantitative_comparision_geometry_aware_image_warping}
\end{figure}

Figure~\ref{fig:quantitative_comparision_geometry_aware_image_warping} summarizes the quantitative results of geometry-aware image warping using different depth inputs. On average, warping with LiDAR depth achieves an SSIM of 0.499 for RGB images and 0.612 for depth images. Using the foundation model-estimated depth improves the SSIM to 0.626 and 0.800, respectively. This corresponds to an improvement of approximately 25.5\% in RGB warping quality and 30.7\% in depth warping quality. Furthermore, we observe that foundation model–based warping remains more robust over longer temporal intervals (i.e., larger frame gaps), whereas LiDAR depth-based warping typically degrades.

The observed improvements mainly come from the foundation model’s ability to generate depth maps with richer structural details and fewer artifacts. In particular, foundation depth estimation better captures object edges, fine surface variations, and subtle geometric discontinuities. LiDAR depth on these regions often becomes sparse or noisy due to limited resolution or reflective materials. These findings highlight the strong potential of foundation model–based depth estimation to improve both the accuracy and robustness of geometry-aware image warping. As a result, we expect the foundation depth model to be used to tackle camera movement and enable accurate cross-frame information reuse. It can also be used to enable high-latency sensing and perception algorithms in real-time applications by allowing the reuse of estimation results on temporally adjacent frames.

\para{Summary}. Foundation models significantly improve the geometry-aware image warping through more accurate depth estimation results. This brings a new opportunity to enable cross-frame information reuse and integrate sparse sensing in real-time mobile AR applications.

\section{Long-Duration Sparse Sensing}
\label{sec:long_duration_sparse_sensing}

Our previous experiment has shown promising results on employing a foundation model-based technique for sparse sensing of temporally adjacent frames. Next, we investigate the scalability of foundation-model-based sparse sensing in the context of 3D environment reconstruction tasks.

\begin{figure}[t]
\centering
    \begin{minipage}{\linewidth}
        \centering
        \includegraphics[width=.8\linewidth,trim=0.5in 3.5in 0in 0in,clip,page=1]{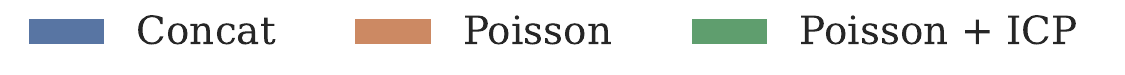}
    \end{minipage}
    \vspace{-2em}

    \begin{subfigure}[t]{0.285\columnwidth}
        \vspace{0em}
        \centering
        \includegraphics[width=\linewidth]{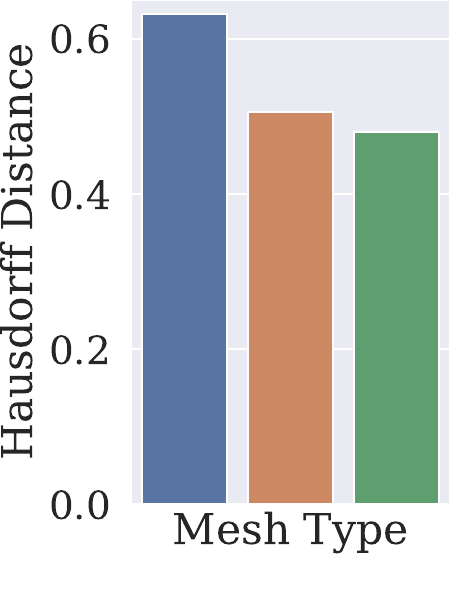}
        \caption{LiDAR}
        \label{subfig:3d_recon_acc_lidar}
    \end{subfigure}\hspace{.05em}
    \begin{subfigure}[t]{0.64\columnwidth}
        \vspace{0em}
        \centering
        \includegraphics[width=\linewidth,trim=0.5in 0in 0in 0in,clip]{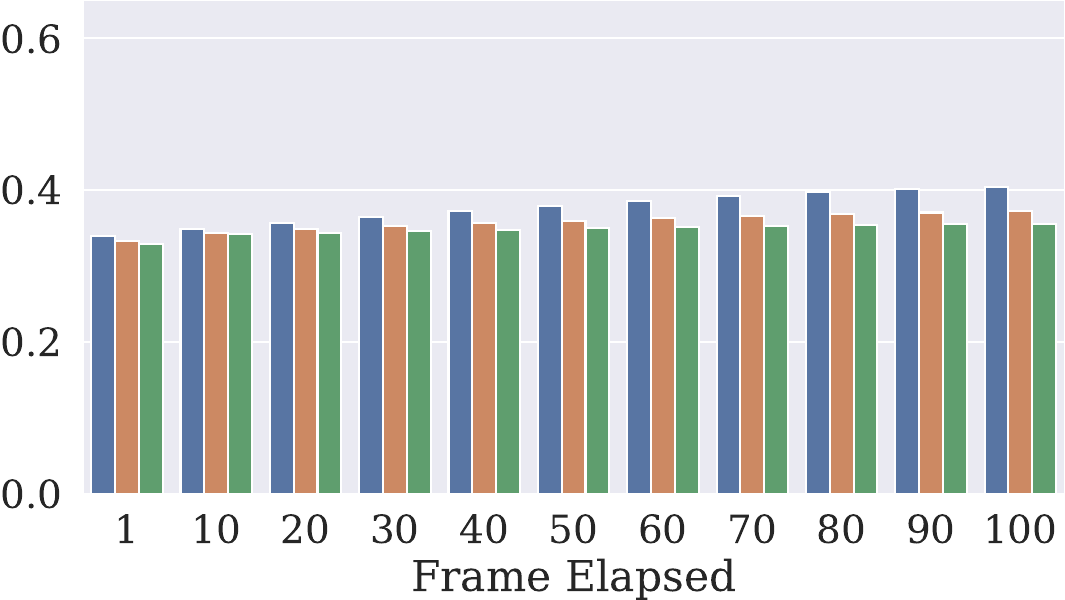}
        \caption{Foudation Model}
        \label{subfig:3d_recon_acc_fm}
    \end{subfigure}
    \vspace{-1em}
    \caption{
        3D reconstruction quality measurement.
        \textnormal{We reconstruct 3D environment meshes with both LiDAR and foundation model-estimated depth and merge multi-view meshes using three different methods. Overall, foundation model-based reconstruction significantly outperforms LiDAR-based methods in terms of Hausdorff Distance$\downarrow$, even under sparse frame inputs.}
    }
    \vspace{-1em}
    \label{fig:3d_recon_acc}
\end{figure}

We evaluate the quality of reconstructing 3D environment meshes from long-duration AR session data. Using the session data from selected scenes, we first reconstruct 3D environment meshes using the device LiDAR depth. The LiDAR depth is from the dense AR frames. We reconstruct the environment mesh for each frame and merge them with three different methods: \1 simple concatenation, \2 Poisson surface reconstruction~\cite{kazhdan2006poisson}, and \3 Poisson surface reconstruction combined with iterative closest point (ICP)~\cite{besl1992method} optimization. Next, we reconstruct the environment mesh using the foundation model-estimated depth on temporally downsampled frames. Similar to our previous experiment, we chose the frame gap from $[0, 100]$ with a step of 10.

Figure~\ref{fig:3d_recon_acc} presents the average reconstruction accuracy of environment meshes across ten scenes using different mesh generation methods.
We use Hausdorff Distance~\cite{huttenlocher2002comparing}, which measures the distance between two subsets of the same metric space, to quantify the reconstruction accuracy. Specifically, in our case, we treat the overlapping region between the reconstructed and ground-truth meshes as the metric space.
As shown in Figure~\ref{subfig:3d_recon_acc_lidar}, applying advanced mesh merging algorithms leads to noticeable improvements in reconstruction quality for LiDAR depth-based methods. However, due to the inherent limitations in the quality of LiDAR depth, the overall reconstruction quality remains low. In contrast, the foundation model-based reconstruction can achieve significantly higher accuracy without advanced merging. Specifically, without temporal downsampling, the foundation depth-based Poisson + ICP reconstruction achieves a Hausdorff Distance of 0.25, whereas the LiDAR depth–based reconstruction yields 0.48. Moreover, we observe that the foundation model–based reconstruction remains robust under sparse input conditions. In Figure~\ref{subfig:3d_recon_acc_fm}, we notice consistently more accurate results even with significantly fewer inputs than the LiDAR-based reconstruction. These findings highlight the strong potential of foundation model–driven sparse sensing for enabling scalable 3D reconstruction tasks.

\para{Summary}.
Our study suggests that foundation models can be leveraged to achieve better 3D reconstruction quality when using sparse sensing compared to directly using LiDAR-based per-frame depth information.

\begin{figure}[t]
\centering
    \begin{subfigure}[b]{0.485\columnwidth}
        \centering
        \includegraphics[width=\linewidth]{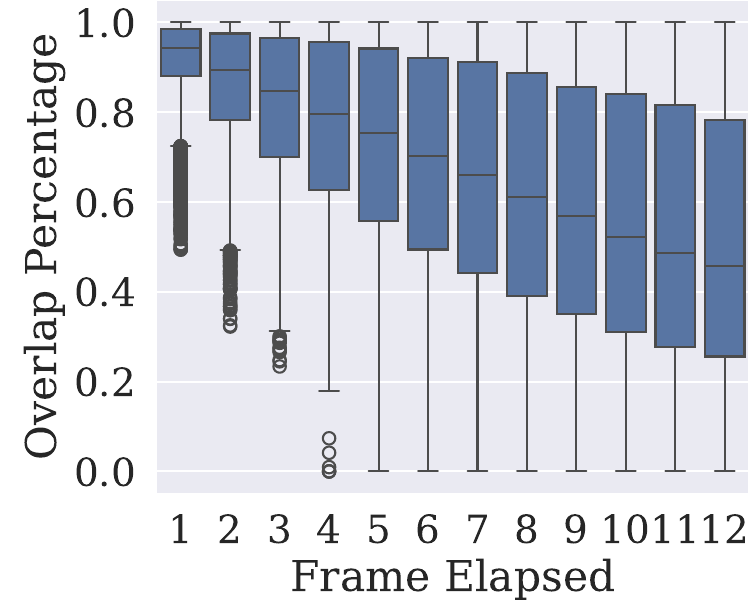}
        \caption{Temporal}
        \label{subfig:overlap_measure_temporal_downsample}
    \end{subfigure}\hfill
    \begin{subfigure}[b]{0.485\columnwidth}
        \centering
        \includegraphics[width=\linewidth]{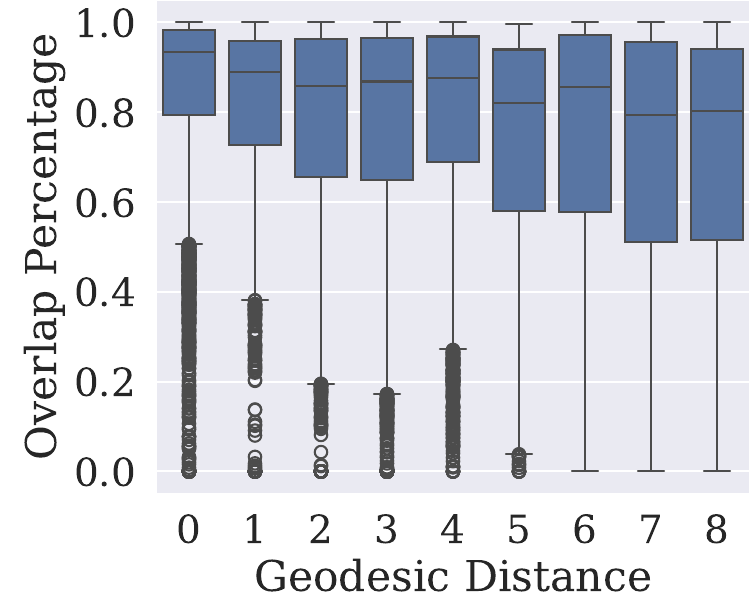}
        \caption{Spatial}
        \label{subfig:overlap_measure_spatial_downsample}
    \end{subfigure}
    \vspace{-1em}
    \caption{
        Measurement of frame overlaps.
        \textnormal{We measure the frame overlaps by calculating the overlap percentage$\uparrow$ of warped pixels between sparse frames. The frames are selected with two policies: time interval-based (a) and geodetic distance-based (b). We observe nonlinearity on information sparsity across both temporal and spatial domains.}
    }
    \vspace{-1em}
    \label{fig:overlap_measurement}
\end{figure}

\section{Towards Spatial-Temporal Sparse Sensing}
\label{sec:spatial_temporal_sparse_sensing}

Although sparse sensing demonstrates great potential for both real-time and long-duration tasks, its effectiveness depends on well-designed control policies that ensure critical information is not missed during the sensing process.
In this experiment, we analyze how information sparsity evolves under different sparse sensing control policies and explore the open questions and challenges for designing such policies. We quantify information sparsity by applying the \textit{geometry-aware image warping} technique and measure the ratio of warped pixels. Different from motion-based analysis, this ratio captures both viewpoint changes and geometric variations in the environment.

We evaluate two categories of sparse sensing policies: \1 \textit{temporal sparse sensing}, which reduces the sensing based on time intervals, and \2 \textit{spatial sparse sensing}, which reduces sensing based on device motion. For temporal sparsity, we vary the inter-frame interval to analyze how decreasing the framerate impacts information overlap. For spatial sparsity, we control camera motion based on the \textit{SE(3) geodesic distance}. This metric, commonly used in 3D vision and SLAM~\cite{triggs1999bundle,gao2023fully,govindu2004lie}, jointly accounts for both rotational and translational differences between camera poses.

Figure~\ref{fig:overlap_measurement} shows our measurement results.
For \textit{temporal sparse sensing}, the information overlap degrades rapidly as the inter-frame interval increases. For example, maintaining an 80\% information overlap requires at most four frames. This finding suggests that a 60~FPS AR stream can be temporally downsampled to 15~FPS, resulting in a 75\% reduction in sensing workload, while still preserving sufficient inter-frame overlap for effective information reuse. A similar trend is observed under spatial sparse sensing. Combined with the geometry-aware image warping technique, this reduction opens up new opportunities for integrating foundation models into real-time AR pipelines at lower frame rates without compromising perceptual consistency.

However, it is important to note that neither temporal nor spatial control policies can strictly guarantee a minimum view-to-view overlap. This is because the information overlap is inherently influenced by user motion dynamics and environmental geometry. Furthermore, our measurements indicate that information sparsity evolves in a nonlinear manner, suggesting that static or heuristic policies may be insufficient for optimal performance. We believe future work should explore hybrid sparse sensing controllers that adapt to both user and environment context to allow intelligent reuse of information across frames to achieve high-quality sparse sensing in mobile AR.

\para{Summary}.
Both temporal and spatial sparse sensing demonstrate high view overlaps, suggesting the promise of exploiting both domains when designing sparse sensing policies.

\vspace{-1em}
\section{Conclusion}
We take a first step toward a foundation model-driven sparse sensing for mobile AR systems. Through our study using real-world AR data, we showed that foundation models can compensate for reduced sensing frequency by more effectively reusing information across temporal and spatial domains, even improving 3D reconstruction quality under very sparsely sensed data.
Our observations that downstream tasks can reuse information from both temporal and spatial domains in AR with the help of foundation models point toward a new class of sensing policy design, i.e., hybrid policies that truly adapt to user environment context and allow mobile devices to sense only when it matters.
\vspace{-1em}

\bibliographystyle{ACM-Reference-Format}
\bibliography{main}

\end{document}